\documentclass{article}

\usepackage[preprint]{neurips_2023}

\usepackage[utf8]{inputenc} % allow utf-8 input
\usepackage[T1]{fontenc}    % use 8-bit T1 fonts
\usepackage{hyperref}       % hyperlinks
\usepackage{url}            % simple URL typesetting
\usepackage{booktabs}       % professional-quality tables
\usepackage{amsfonts}       % blackboard math symbols
\usepackage{nicefrac}       % compact symbols for 1/2, etc.
\usepackage{microtype}      % microtypography
\usepackage[ruled,linesnumbered,vlined]{algorithm2e}
\usepackage{graphicx}
\usepackage{xcolor}
\usepackage{amsmath}
\usepackage{ulem}

\newcommand{\grad}{\nabla}
\newcommand{\be}{\begin{equation}}
\newcommand{\ee}{\end{equation}}
\newcommand{\cE}{\mathbb{E}}

\title{Stochastic Ratios Tracking Algorithm for Large Scale Machine Learning Problems }

\author{%
  Shigeng~Sun \\
  Department of Engineering Sciences and Applied Mathematics\\
  Northwestern University\\
  Evanston, IL, USA \\
  \texttt{shigengsun2024@u.northwestern.edu} \\
   \And
  Yuchen~Xie \\
  \texttt{ycxie@u.northwestern.edu} \\
}

\begin{document}

\maketitle

\begin{abstract}
Many machine learning applications and tasks rely on the stochastic gradient descent (SGD) algorithm and its variants. Effective step length selection is crucial for the success of these algorithms, which has motivated the development of algorithms such as ADAM or AdaGrad. In this paper, we propose a novel algorithm for adaptive step length selection in the classical SGD framework, which can be readily adapted to other stochastic algorithms. Our proposed algorithm is inspired by traditional nonlinear optimization techniques and is supported by analytical findings. We show that under reasonable conditions, the algorithm produces step lengths in line with well-established theoretical requirements, and generates iterates that converge to a stationary neighborhood of a solution in expectation. We test the proposed algorithm on logistic regressions and deep neural networks and demonstrate that the algorithm can generate step lengths comparable to the best step length obtained from manual tuning.
\end{abstract}

\section{Introduction}

Optimization {problems that involve millions of unknown parameters} and vast datasets  are common occurrences in machine learning. Addressing the computational demands of such problems necessitates highly efficient implementations of stochastic gradient methods \cite{Goodfellow-et-al-2016-Book,bottou2018optimization}. In this context, the paper proposes an algorithm that adaptively tunes the step length (or learning rate), striving to account for the presence of nonlinearity in the true objective function and stochasticity in the function and gradient approximations used in the iteration.

This algorithm is designed to minimize empirical risk,
\be
F(w)=\frac{1}{n} \sum_{i=1}^n f\left(w ; \xi_i\right) := \frac{1}{n} \sum_{i=1}^n f_i(w), 
\ee
where $\left(\xi_i\right)_{i=1}^n$ denote the training examples and $f(\cdot ; \xi): \mathbb{R}^d \rightarrow \mathbb{R}$ is the composition of a prediction function (parametrized by $w$ ) and a loss function. 
$\{\xi_k\}$ can be seen as representing a sequence of jointly independent random variables. The training problem consists of finding an optimal choice of the parameters $w \in \mathbb{R}^d$ with respect to $F$, i.e.,
\be
\min _{w \in \mathbb{R}^d} F(w)=\frac{1}{n} \sum_{i=1}^n f_i(w)
\ee
For these type of problems, the current \textit{de facto} optimization methods are the stochastic gradient descent (SGD) method and its variants, such as Adam and AdaGrad. More concretely, a basic mini-batch stochastic gradient algorithm at iteration $k$ involves the update %\shig{[I thought that the definition of $\alpha_k$ may be ambiguous. We can make up our minds by using it to denote either the trial step or the actual update.]}
\be w_{k+1} = w_k - {\alpha_k} \grad  F_{S_k}\ee
where we define
\be
\nabla  F_{S_k} \left(w_k\right)=\frac{1}{\left|S_k\right|} \sum_{i \in S_k} \nabla f_i\left(w_k\right)
\ee
and the set $S_k \subset\{1,2, \ldots\}$ indexes training data points. The sample $S_k$ changes at every iteration and in the basic mini-batch stochastic algorithm, its cardinality $|S_k|$ remains the same.
The step size parameter of SGD algorithm is is often difficult to determine and a large number of approaches have been proposed to address it, and none has been universally adopted.

The recent paper by Defazio et al. \cite{defazio2023learning} proposes a steplength technique based on worst-case complexity. It requires the estimation of the distance $D$ from the initial point to the solution set as well as the estimation of the Lipschitz constant $G$. In this paper, we take the view that the step length parameter should depend on $G$ but also on the variance in the stochastic gradient approximations (as opposed to the distance to the solution).

\paragraph{Our Goals} Whilst we do not purport that our algorithm is designed with the primary objective of surpassing some of the existing state-of-the-art computational algorithms \cite{kingma2014adam,duchi2011adaptive,babanezhad2015stop,NIPS2014_5258} in terms of efficiency or speed, our intention is to explicate the principles that govern the selection of step lengths for stochastic algorithms in the most interpretable and transparent fashion. We have also endeavored to demonstrate the intricacies of such an algorithm that incorporates these principles. Furthermore, we contend that by allowing the discretion to select {search} directions $d_k$ (as expounded in the algorithmic section), this framework can be adapted to various algorithms, e.g. \cite{gadat2018stochastic,hu2009accelerated}, not just the standard SGD algorithm. %and iterate directions, and thus, discussions on these algorithms will be deferred to future endeavors in the interest of lucidity.

\paragraph{Stochastic Line Search or Stochastic Trust Region Methods} 

%\cite{sun2022trust,kingma2014adam,mybook,jin2021high,dangel2020backpack,bottou2018optimization,babanezhad2015stop,SW96,shi2022noise,RobMon51,Goodfellow-et-al-2016-Book,NIPS2014_5258,byrd2012sample,berahas2016multi} cifar10\cite{krizhevsky2009learning}GISETTE\cite{guyon2004result}

%
%In conventional non-stochastic optimization settings, strategies such as trust region and line search were introduced to regulate step lengths as the direction in proximity to a current iterate is typically a descent direction \cite{mybook}. In contrast, in the stochastic setting, the quality of the direction is sometimes questionable and whether an algorithm should spend more effort exploring that direction remains debatable. Nevertheless, rather than regarding step length selection as an external input that necessitates sophisticated tuning, numerous algorithms have been devised based on classical nonlinear optimization techniques to determine the step length during runtime. 

In classical optimization, techniques like trust region and line search are used to control step lengths since near the current iterate, the step direction generated is usually a descent direction\cite{mybook}. However, in stochastic optimization, the direction's quality can be questionable, and the value of exploring it is debated. Despite this, instead of relying on external inputs for step length selection, several algorithms use classical nonlinear optimization methods to dynamically determine step lengths during runtime. For instance, \cite{cartis2015global,paquette,shi2022noise} have each examined stochastic line search algorithms that potentially involve multiple function evaluations on a given direction; \cite{jin2021high} introduced a stochastic step search algorithm that modifies the search direction each time a step is refused under a relaxed Armijo condition; in terms of trust region approaches, \cite{toint2021TRqEDAN,curtis2019stochTR,sun2022trust} have each proposed remedies for adapting existing algorithms to stochasticities.
%In \cite{sun2022trust}, a trust region approach was adopted, which presumes that the imprecisions in the evaluations are restricted. 

In stead of treating step lengths as exogenous quantities that may require substantial tuning or using a pre-set diminishing step length as proposed in \cite{RobMon51}, we propose to employ a progress ratio to gauge the average progress made in past iterations and differentiate the effect of noise into a separate ratio. This obviates the potential additional computation for the algorithm to backtrack multiple times before taking a step and allows for a more comprehensive assessment of the overall progress made in the previous iterations.

%\paragraph{Nano-batching} Stochastic Nano-Batching Algorithm?

%[here discuss backpack]

\paragraph{Contributions} We propose a novel algorithm for determining the appropriate step length at the runtime, thus avoiding extensive tuning. The numerical experiments show that the method proposed in this paper---which we call the Stochastic Ratios Tracking method---produces step lengths that are comparable to the best hand-tuned step lengths while achieving a good balance between computational costs and training efficiency. Analytical results show convergence in expectation of the algorithm under mild conditions while highlighting the effect of nonlinearity and stochasticity during the design of step length selection principles.

\section{The Stochastic Ratio Tracking Algorithm}

When tackling stochastic optimization problems in machine learning, selecting an appropriate step length parameter for gradient-based first-order methods depends on two key factors: nonlinearity and stochasticity. {Nonlinearity, which is sometimes represented as curvature information, determines the optimal choice of step length in a given direction {in the conventional non-stochastic settings}.} Stochasticity, on the other hand, relates to inaccuracies or noise in gradient estimates, and is often characterized by variances or second moments. To achieve optimal performance, {one must take into consideration both non-linearity and stochasticity.} 

%\sout{one might choose longer step lengths that result in satisfactory progress in terms of nonlinearity, while also using smaller step lengths for noisy gradient estimations to limit damage}

%Our specific aim is to introduce two ratios that serve to characterize nonlinearity and stochasticity, respectively. 

%These ratios enable the separation and identification of the distinct effects of nonlinearity and stochasticity. Extensive literature has established the significance of these two aspects, and we leverage these classical findings to underpin the algorithm described in this section. 

To motivate the proposed procedure, we cite a well-known result about the stochastic gradient method (c.f. \cite{bottou2018optimization}, Theorem 4.6). It has been established that if the step length is given by 
\be\alpha = \frac{1}{L (M_V+1)},\label{step_choice}\ee 
(where $L$ is the Lipschitz constant, and $M_V$ is a scalar defined below) then the iterates given by the stochastic gradient algorithm converge to a stationary point in expectation under certain conditions. Specifically, the iterations update $w_{k+1} = w_k - \alpha \grad  F_{S_k}$ yields 
\be\mathbb{E}\left[F\left(w_k\right)-F_*\right] \leq \left(1- \frac{1}{L (M_V+1) }\right)^{k-1}\left(F\left(w_1\right)-F_*\right),\ee 
where $F^*$ is the optimal function value. A result of this type can be established under various conditions (c.f. \cite{bottou2018optimization}, Assumptions 4.1 to 4.3); one of the conditions is 
\be \cE_{\xi_k}[\| \nabla f\left(w_k, \xi_k\right) \|^2]-\|\nabla F(w_k)\|^2\leq M_V\|\nabla F(w_k)\|^2, \label{eq:Mv_no_M}\ee
%\yx{([YX]: without the constant term $M$ in BCN2018 Assumption 4.3, this assumption is very strong and usually not satisfied.)}. 
here $F(w_k)$ denotes the true objective. We can approximate this condition at each given iterate $k$ as:
%\be \frac{\frac1{|S_k|}\sum_{i \in S_k}[\| \nabla f_i\left(w_k\right) \|^2]-\|\nabla   F_{S_k}(w_k)\|^2}{\|\nabla  F_{S_k}(w_k)\|^2}\leq M_V. \ee
\be \frac{\sum_{i \in S_k}\| \nabla f_i\left(w_k\right) \|^2 {/|S_k|} -\|\nabla   F_{S_k}(w_k)\|^2}{\|\nabla  F_{S_k}(w_k)\|^2}\leq M_V. \label{eq:samp_Mv}\ee
%\yx{([YX]: there are issues with notations here. Suggest that we use summation instead of expectation here.)}
We can compute the value on the left-hand side using Backpack \cite{dangel2020backpack} within an iteration, which provides an approximation of $M_V$, that we shall name the \textit{variance ratio estimate} later in this paper.

 While many problems in machine learning satisfy (\ref{eq:Mv_no_M}) \footnote{This condition is sometimes referred to as `homogeneity of minima': a minimizer of $F$ is a minimizer of $f$ with probability one \cite{patel2017impact}.}, and in particular in some over-parametrized models in data science \cite{bassily2018exponential}, this assumption still can be perceived as somewhat strong. A more general condition is 
 $$\cE_{\xi_k}[\| \nabla f\left(w_k, \xi_k\right) \|^2]-\|\nabla F(w_k)\|^2\leq M_V\|\nabla F(w_k)\|^2 + M.$$
We will discuss later that the presence of a constant term $M \geq 0$ on the right-hand side would only {potentially} introduce an overestimation of $M_V$ in our algorithm, which doesn't affect convergence, as the size of the overestimation is bounded above outside of a neighborhood of stationarity. 
%\yx{\sout{without a constant term $M$ added to the right-hand side of (\ref{eq:Mv_no_M})},}

%\jn{With estimates [of $L$] and $M_V$ at hand, one can define the step length by (\ref{step_choice}); but since these estimates are not accurate, the value of $\alpha$ will be adjusted periodically using a progress ratio test described next. }

Besides using the noise ratio $M_V$ to account for stochasticity, we shall employ another ratio estimate (\textit{progress ratio estimate}) to allow the step to be adjusted to scale with $\frac1L$ as suggested in (\ref{step_choice}). In the subsequent section, we shall explain the computations of the two ratios above. %which is based on the progress made in previous updates, 

%To motivate the proposed procedure, we cite the following fundamental lemma from BCN 2018: (Assumption 4.1 to 4.3, Theorem 4.6, with unbiased gradient and a choice of step length adaptation for technical convenience)

%\paragraph{Fundamental Lemma} \textit{
%Assume that $F$ is $\mu, L$ smooth in $\omega$ and $\nabla f_k$ follow 
%$\cE_{\xi_k}[\nabla f\left(w_k, \xi_k\right)] = \nabla F_k,$ and
%$\cE_{\xi_k}[\| \nabla f\left(w_k, \xi_k\right) \|^2]-\|\nabla F_k\|^2\leq M_V\|\nabla F_k\|^2\shig{+M}$
%then running fixed step length 
%\be\alpha = \frac1{LM_V}\ee
%with iterate updates $w_{k+1} = w_k - \alpha \grad f_k$ yield 
%\be\mathbb{E}\left[F\left(w_k\right)-F_*\right] \leq \left(1- \frac{1}{L M_V}\right)^{k-1}\left(F\left(w_1\right)-F_*\right)+ O(M)\ee
%}

%This lemma highlights the impact of nonlinearity and stochasticity ($L$ and $M_V$, respectively) on an efficient step size. Developing reliable techniques for obtaining direct or indirect estimates of these quantities constitutes a critical component of our algorithm. The remaining part of this section is organized so as to introduce the techniques for these tasks and formulate the algorithm.

\subsection{Step Length Selection}

\paragraph{Progress Ratio Estimates} 
%As explicated in the preceding section, classical optimization techniques frequently resort to line search or trust region methodologies to regulate the step length and guarantee adequate advancement towards stationarity. Nonetheless, in the case of a noisy computed direction, the implementation of such algorithms may pose difficulties. While prior studies have addressed this issue in \cite{albertls,shigengtr}, the utilization of noisy directions to attain suitable step lengths is still a topic of debate, particularly in the presence of substantial noise. 
{ In this work, we abstain from employing back-tracking line search or shrinking trust region within an iteration; instead, we opt to calculate a ``Progress Ratio Estimate'' to evaluate the effectiveness of the current step length along the intended direction. After sufficient information about past iterations is attained, we update the step length accordingly. }

%\shig{[be more careful about notation] 

%$$
%w_{k+1} \leftarrow w_k-\alpha_k \nabla f_{i_k}\left(w_k\right) .
%$$
%Here, for all $k \in \mathbb{N}:=\{1,2, \ldots\}$, the index $i_k$ (corresponding to the seed $\xi_{\left[i_k\right]}$, i.e., the sample pair $\left.\left(x_{i_k}, y_{i_k}\right)\right)$ is chosen randomly from $\{1, \ldots, n\}$ and $\alpha_k$ is a positive stepsize. Each iteration of this method is thus very cheap, involving only the computation of the gradient $\nabla f_{i_k}\left(w_k\right)$ corresponding to one sample. The method is notable in that the iterate sequence is not determined uniquely by the
%}

More rigorously, within an iteration $k$, we generate an index set $S_k$ and fix it. We compute the stochastic gradient $\grad  F_{S_k}(w_k)$ and the {search} direction $d_k$ (for instance, SGD has $d_k = -\grad  F_{S_k}(w_k)$; this framework also allows other choices of $d_k$). We then evaluate the \textit{Progress Ratio Estimate} with step length $\alpha$ as $\hat \rho_k(\alpha)$:
\be\hat \rho_k(\alpha) = \frac{ F_{S_k}( w_k + \alpha d_k) -  F_{S_k}(w_k)}{\alpha \grad  F_{S_k}^T(w_k)d_k }.\label{eq:hatrho}\ee

{Here the same sample $S_k$ is used in \eqref{eq:hatrho}. Note that the condition $\rho_k(\alpha)\geq c_1$ is the Armijo condition on $F_{S_k}$ stating that the step length is to have provided a sufficient decrease in the objective. We track ratio (\ref{eq:hatrho}) over a set of iterations and if the average value of $\hat \rho$ in these iterates is large enough, we increase the step length, as this suggests that more progress may be achievable. On the other hand, if that ratio tracks to a small value, we shrink $\alpha$ to facilitate the alignment between progress and expectation. In other words, this non-dimensional ratio $\hat \rho$ is employed to assess the degree of alignment between the stochastic objective function and its local linear model. A value of the ratio in proximity to 1 indicates the stochastic objective function exhibits linear behavior along the direction $d_k$ and the progress achieved by traversing a distance of $\alpha$ in this direction is in accordance with the anticipated progress.

Our progress ratio tracking, therefore, determines whether the steplengths should be adjusted. We now consider how to choose the actual value of $\alpha_k$.}
%While astute readers may note that the `Progress Ratio Estimate' is akin to the ratio employed by trust region algorithms when the function model is linear and that comparison of this ratio to a pre-determined threshold value $c_1$ establishes the standard Armijo condition, 
%we shall expound on several salient properties of this ratio: 
%(i) \textit{Sample Consistency}: the numerator of the quantity requires evaluation of the same stochastic objective $f_k$ at the current iterate $w_k$ and a trial point $w_k + \alpha d_k$, and that the sample $\xi_k$ (hence $f_k(\cdot)$) is fixed during this process; (ii) \textit{Evaluation Cost}: One typically expects that the evaluation of $f_k$ to be relatively non-expensive and can cost as much as evaluating the stochastic gradient. (iii) \textit{Performance Metric}: \shig{}

\paragraph{Variance Ratio Estimates}
{As already mentioned in the paragraph surrounding (\ref{step_choice}), the step length must depend on the error/noise in the gradient approximation, it's natural to measure this noise in the form of the mini-batch variance. Thanks to the recent advent of Backpack \cite{dangel2020backpack}, obtaining estimates of $\hat M_V$ from mini-batch updates has become relatively effortless in PyTorch, with only minor computational overhead. We perform this as follows:}

%The existence of parameter $M_V$ in the fundamental lemma in the preceding section was probably introduced from the very beginning stage of understanding and analysis of the stochastic gradient descent algorithm and remains largely of theoretical interest thus far to the best of our knowledge. Due to the computational complexity and challenges in devising effective procedures, the estimation of $M_V$ has been largely absent in contemporary literature for practical usage. Nevertheless, with the advent of Backpack \cite{backpack}, obtaining estimates of $M_V$ from mini-batch updates has become relatively effortless, with only minor computational overhead.

We re-iterate that the stochastic objective $F_{S_k}$ is often referred to as a `mini-batch' with cardinality $|S_k| = m$ (where $m = 16,32,64,...$ are commonly used):
\be
F_{S_k} \left(w_k\right)=\frac{1}{\left|S_k\right|} \sum_{i \in S_k}  f_i\left(w_k\right) 
,\quad \nabla  F_{S_k} \left(w_k\right)=\frac{1}{\left|S_k\right|} \sum_{i \in S_k} \nabla f_i\left(w_k\right)
\label{eq:gradF}
\ee
{This motivates an estimated empirical value $\hat M_{V,k}$, that would allow the satisfaction of (\ref{eq:samp_Mv}) when used in place of $M_V$}:
\be
\hat M_{V,k} =\frac{\hat {V}_k }{ \|\grad F_{S_k}(w_k)\|^2},
\quad \text{with
} \quad 
%\hat {V}_k= \frac{ \sum_{i =1}^m \|\nabla f_i\left(w_k\right)\|^2- \| \grad F_{S_k}(w_k) \|^2}{|S_k|  -1}.
\hat {V}_k =\sum_{i \in S_k}\| \nabla f_i\left(w_k\right) \|^2/ |S_k| -\|\nabla   F_{S_k}(w_k)\|^2
\label{eq:hatM}
\ee
{This dimensionless ratio $\hat M_{V,k}$ is used to quantify the magnitude of noise inherent in the estimates of stochastic gradients. As explained before, it serves as an indicator of the relative second moment of the stochastic gradients, a quantity that an ideal step length should inversely dependent upon.}%, as explained in the paragraph surrounding (\ref{step_choice}).

%{We comment again on the condition in (\ref{eq:Mv_no_M}). A weaker assumption can be written as $\cE_{\xi_k}[\| \nabla f\left(w_k, \xi_k\right) \|^2]-\|\nabla F(w_k)\|^2\leq M_V\|\nabla F(w_k)\|^2+M$.
%To allow for satisfying this assumption, $\hat M_{V,k}$ is the empirical value of $M_V + M/\|\grad F_{k}(w_k)\|$ (instead of $M_V$). Before reaching a solution outside of a stationary neighborhood of size $\epsilon$, i.e. when $\|\grad F_{k}(w_k)\| > \epsilon$, the overestimation $ M/\|\grad F_{k}(w_k)\|$ is bounded above by $M/\epsilon$, and does not affect the convergence of the algorithm.}
{We comment again on the condition in (\ref{eq:Mv_no_M}). A weaker assumption can be written as $\cE_{\xi_k}[\| \nabla f\left(w_k, \xi_k\right) \|^2]-\|\nabla F(w_k)\|^2\leq M_V\|\nabla F(w_k)\|^2+M$. If this weaker assumption is satisfied (with $M > 0$) instead of \eqref{eq:Mv_no_M}, we might overestimate $M_V$, but the overestimation will be no more than $M/\|\grad F_{k}(w_k)\|$, which is bounded above by $M/\epsilon$ if $\|\grad F_{k}(w_k)\| > \epsilon$, i.e., if the iterates are outside the neighborhood of a stationary point. Hence, replacing \eqref{eq:Mv_no_M} with a weaker assumption doesn't affect the convergence of the algorithm to the stationary neighborhood.}

%\shig{[ re-write a bit to make clearer. ] This ratio $\hat M_{V,k}$ serves to quantify for the level of noise present in the stochastic gradient estimates and can be interpreted as a measure of the bound on the second moment of the stochastic gradient.} While the sample variance $\hat{V}_k$ scales with $\|\grad F_{S_k}\|^2$, $\hat M_{V,k}$ is a dimensionless quantity which quantifies the level of noise in the stochastic gradient estimations, and according to the section surrounding (\ref{step_choice}), appropriate step lengths should be inversely proportional to this quantity.

{In the next section, we shall proceed to expound upon the algorithm. It is pertinent to reiterate that the incorporation of the \textit{progress ratio estimates} $\hat \rho_k$ is intended to factor in nonlinearity and enable the scaling of steps with $1/L$, while the \textit{variance ratio estimates} $\hat M_{V,k}$ are intended to {account for} the impact of noise or inaccuracies in the gradient estimates. By isolating and addressing the effects of nonlinearity and noise {separately}, the algorithm {can leverage mechanisms designed for classical, non-stochastic non-linear optimization in the stochastic setting}.
% We note that a nano-batch size of 1 was chosen, which facilitates the estimation of the aforementioned quantity through the use of Backpack\cite{backpack}.

%\paragraph{Lipschitz Estimates}

\subsection{Specification of the Algorithm}

In the $k$-th iteration, upon generating $ \grad F_{S_k}(w_k) $ and $d_k$, our proposed algorithm calculates the quantities $\hat \rho_k(\alpha)$ for a trial step size $\alpha$ and $\hat M_{V,k}$ using the expressions provided in the previous subsection. These values, $\hat\rho_k(\alpha)$ and $\hat M_{V,k}$, are then appended in buffers $v_\rho$ and $v_M$, respectively. (If the size of $v_\rho$ and $v_M$ exceed a predetermined memory length $N$, we delete the oldest values from the buffer.) The algorithm then takes the step:
\be w_{k+1} = w_k + \frac{\alpha_k}{mean(v_M)+1} d_k\label{eq:update}\ee
Once the size of $v_\rho$ reaches the pre-specified memory length $N$, the algorithm decides whether to increase or decrease $\alpha_k$ based on the values in $v_\rho$. Our strategy is to compute the mean of the values in $v_\rho$ (denoted as $\bar \rho$) and compare it with user-defined parameters $c_1$ and $c_2$, where $0<c_1\leq c_2<1$. If $\bar \rho <c_1$, then $\alpha$ is reduced, and if $\bar \rho>c_2$, then $\alpha$ is increased. %\yx{(Can be deleted) One can also develop hypothesis testing to obtain confidence guarantees for the $\rho$ values computed by assuming their underlying distribution. These tests can provide high probability guarantees for convergence. However, we have not observed significant differences in numerical performance when using more advanced statistical tests compared to simple statistical average comparisons. }%Hence, for the sake of being concise and specific, we continue to use statistical average comparisons. Nonetheless, this area is open for future research and possible developments. }

We are now ready to state our algorithm below:\\
\begin{algorithm}[H]\label{algorithm}
\SetAlgoLined
 Initialize $w_0$ and $\alpha_0$, pick $0<c_1\leq c_2<1$, $\tau>1$ and memory buffers $v_\rho$ , $v_{M}$ and length $N$.\\
 \For{$k=0,1,...$}{
Generate index set $S_k$, evaluate $\nabla  F_{S_k}$ by (\ref{eq:gradF}), compute $d_k$;\\
%\be\nabla  F_{S_k} \left(w_k\right)=\frac{1}{\left|S_k\right|} \sum_{i \in S_k} \nabla f_i\left(w_k\right)\ee\\
Evaluate $\hat M_{V,k}$ by (\ref{eq:hatM}) and $\hat \rho_k(\alpha_k)$ by (\ref{eq:hatrho});
%\be\hat M_{V,k} =\frac{\hat \mathbb{V}_k }{ \|\grad F_{S_k}(w_k)\|^2}, \quad \hat \mathbb{V}_k= \frac{ \sum_{i =1}^m \|\nabla f_i\left(w_k\right)\|^2- \| \grad F_{S_k}(w_k) \|^2}{|S_k|  -1}\ee
 \\
%Evaluate \\
%\be\hat \rho_k(\alpha) = \frac{ F_{S_k}( w_k + \alpha d_k) -  F_{S_k}(w_k)}{\alpha \grad  F_{S_k}^T(w_k)d_k }.\ee\\
Append $ \hat \rho_k(\alpha_k)$ to $v_\rho$ and $\hat M_{V,k}$ to and $v_{M}$; if size of a buffer exceeds $N$, delete oldest element;\\
 \If{size of $v_\rho=N$}{
 	\uIf{$mean(v_\rho)>c_2$}{$\alpha_{k+1}=\tau\alpha_k$\\ clear $v_\rho$}\ElseIf{$mean(v_\rho)<c_1$}{$\alpha_{k+1}=\alpha_k/\tau$\\ clear $v_\rho$}}
 Update iterate by (\ref{eq:update});%\be w^{k+1} = w_k + \frac{\alpha}{{mean(v_M)}} d_k\ee\\
 % Set $k \leftarrow k+1$;
 }
\caption{Stochastic Ratio Tracking Algorithm (SRT)}
\end{algorithm}
To illustrate the implementation of this algorithm, we provide the following schematics in Figure \ref{fig:sche}.
\begin{figure}[h!]\label{fig:sche}
    \centering
    \includegraphics[width = 0.4\textwidth]{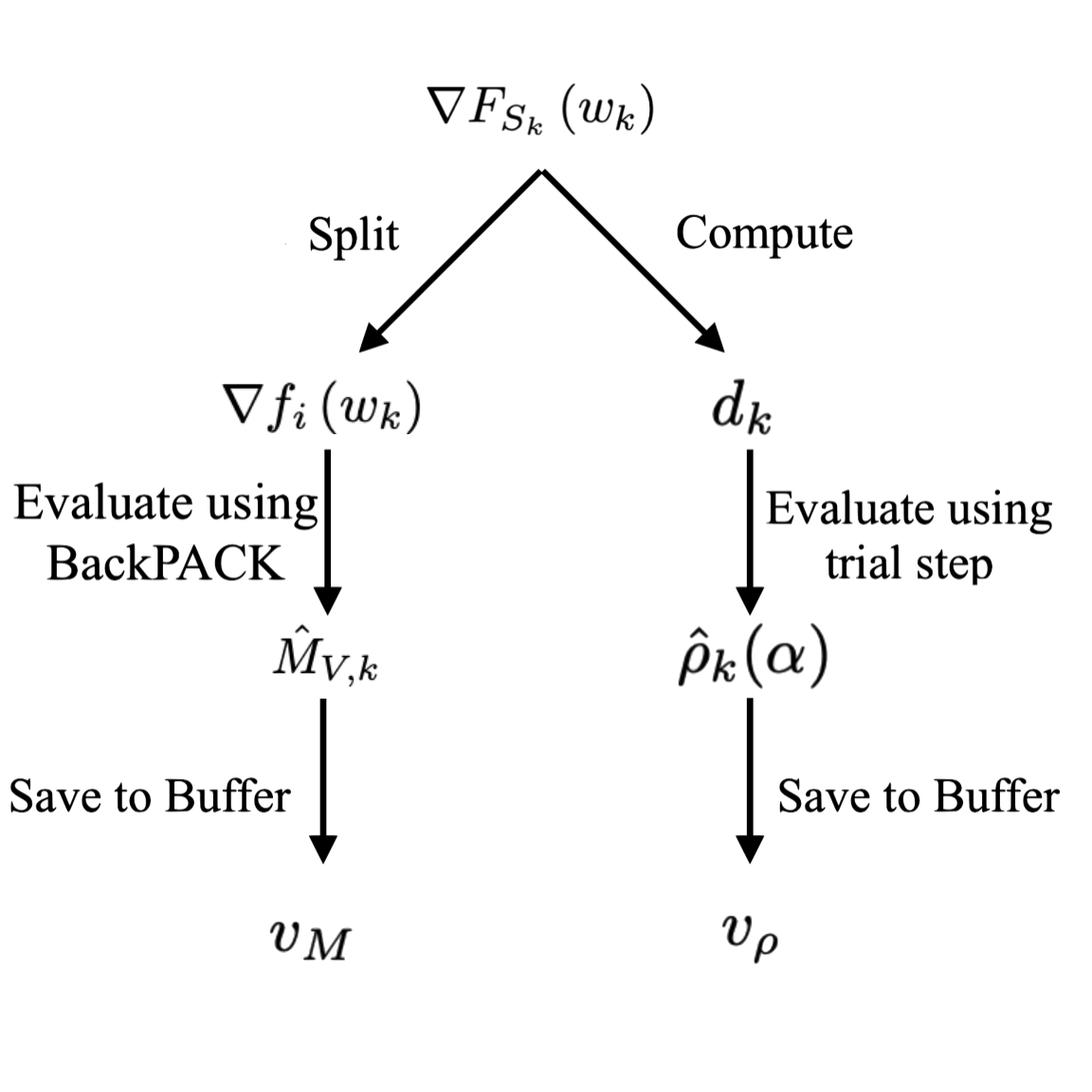}
    \includegraphics[width = 0.4\textwidth]{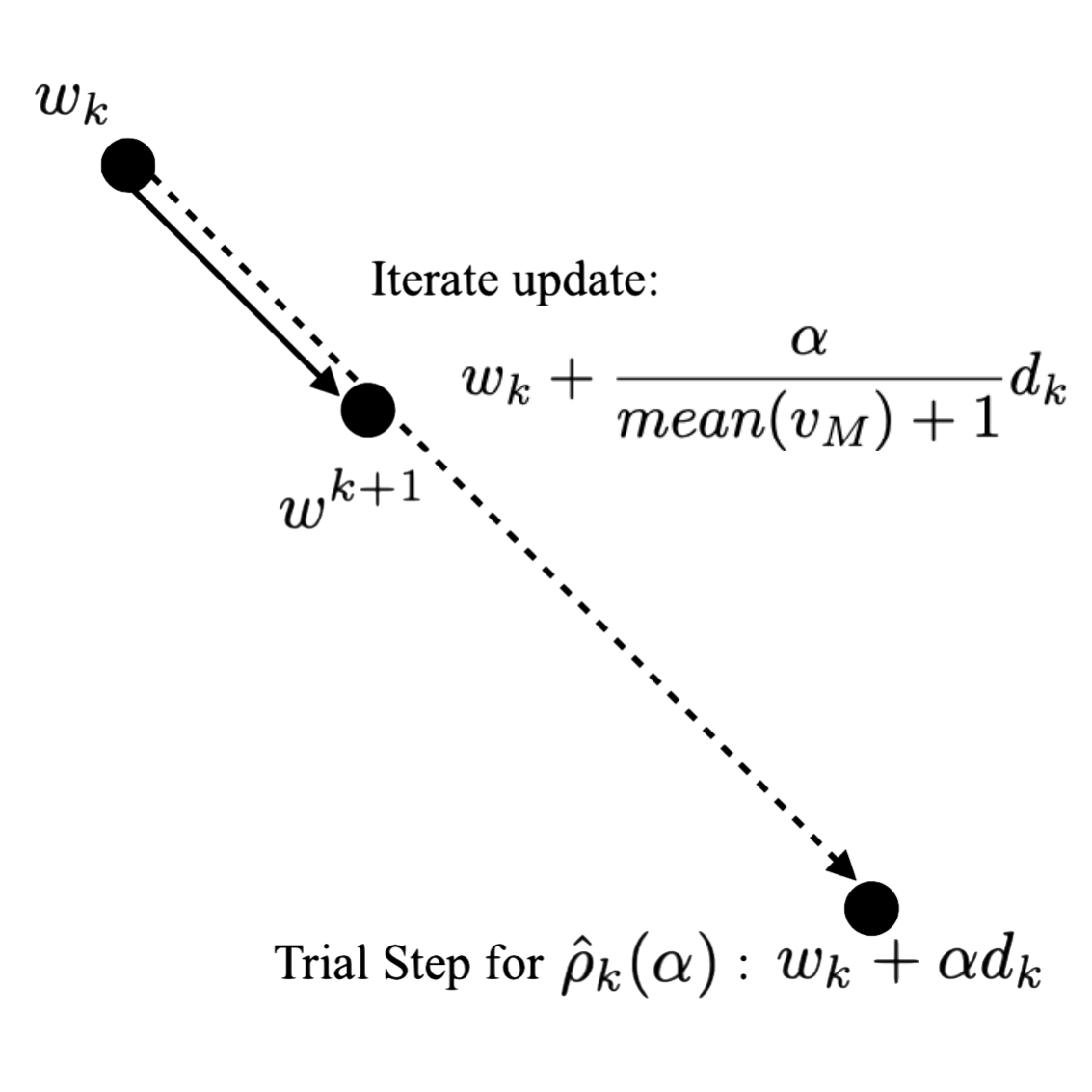}
    \caption{Schematics of the SRT algorithm. Left: computation of estimates. Right: Trial step and iteration updates.}
    \label{fig:sche}
\end{figure}

%\yx{(Can be deleted) Prior to presenting convergence theory and numerical performance of SRT {on stochastic optimization problems}, we provide numerical experiments (using constant step lengths over fixed batch sizes) in the next section to demonstrate the estimated values of $\hat M_V$ and $\hat \rho(\alpha)$.}

\section{Fixed Step Lengths and Batch Sizes Experiments}

We first showcase the ratios $\hat \rho_k, \hat M_{V,k}$ in controlled settings before presenting convergence theory. More specifically, for Logistic regression and DNN problems, we employ the stochastic gradient algorithm with different fixed step lengths and across different batch sizes. We show that the choice of step length $\alpha$ has a direct impact on the estimated and $\hat \rho_k$, and similarly the choice of batch size $\|S_k\|$ affects the estimated and $\hat M_{V,k}$ as expected. %In this fashion we obtain numerical results which supports theory predictions.

%$d_k = -\nabla F_{S_k} \left(w_k\right)$, with different constant step lengths and batch sizes across diverse datasets and utilizing logistic regression and deep neural networks.

%The initial series of experiments pertains solving logistic regression problem on the gisette dataset \cite{gisette} with fixed step length stochastic gradient algorithm. The outcomes of the experiments are presented in the form of two typical result sets in the two panels of Figure \ref{logi_batch}, wherein a fixed step length stochastic gradient algorithm (for 10 epochs) is implemented with batch sizes of 32 and 128 on the left and right panels, respectively.

\subsection{Batch size's influence on $\hat M_{V,k}$}

As discussed in the previous sections, the values of $\hat M_{V,k}$ quantify the level of stochasticities in the gradient estimations, where a better estimation of the gradient in a larger-sized batch should be reflected in smaller values in the estimations of $\hat M_{V,k}$. We demonstrate this effect in this subsection.

We start the initial round of experiments in Figure \ref{fig:logi_batch} with logistic regression on the gisette dataset \cite{guyon2004result}, where we run the SGD algorithm with a constant step length of 0.003 for 10 epochs. We repeat the same experiment with batch size $\|S_k\|$ of $8$ and $64$, and included, in each of the left and right panels, an example of typical results in Figure \ref{fig:logi_batch}.  In each of the panels, from top to bottom, we report $F_{S_k}(w_k)$, step length, $\hat M_{V,k}$ and $\hat \rho_k$, respectively.
%The results from running the fixed step length stochastic gradient algorithm (for 10 epochs) with batch sizes of 32 and 128 are presented in the left and right panels in Figure \ref{fig:logi_batch}, respectively, as indicative of typical outcomes.
\begin{figure}[h!]
    \centering
    \includegraphics[width = .45\textwidth]{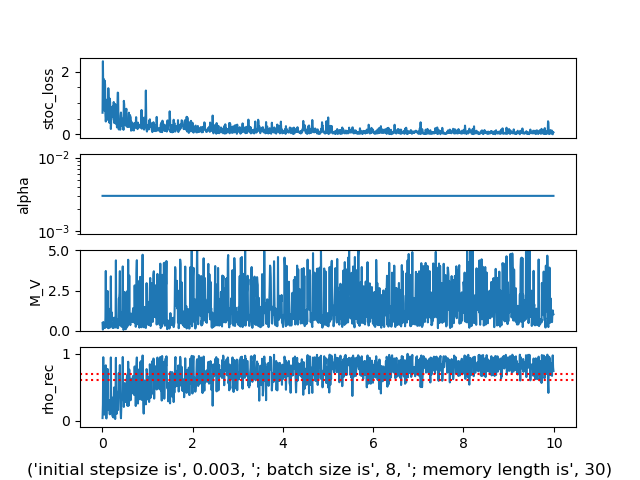}
    \includegraphics[width = .45\textwidth]{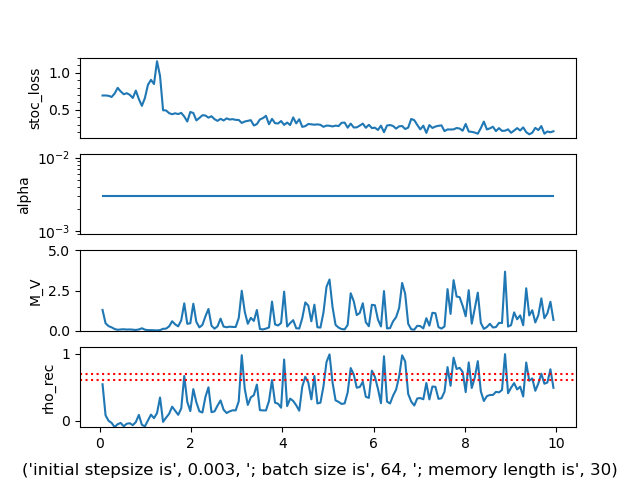}
    \caption{Logistic regression with the gisette data. Each of left and right panels contains a typical run with a constant step length of 0.003; each panel contains from top to bottom: $F_{S_k}(w_k)$, $\hat M_{V,k}$, $\mathbb{V}_k$, $\hat \rho_k(\alpha)$; the left panel has a minibatch size of $8$ while the right panel has a minibatch size of $64$.}
    \label{fig:logi_batch}
\end{figure}

As we observe from Figure \ref{fig:logi_batch}, throughout the training progress, the values of $\hat \rho_k$ remained largely comparable when batch size increased from 8 to 64; however, $\hat M_V$ values reduced, indicating that the noise in the gradient is reduced and larger steps can be favored. 

We continue to perform similar experiments on Deep Neural Networks. We train a 3-layer feed-forward neural network on the Fashion-MNIST data, using a constant step length of $0.5$. Similar to before, we vary the batch size from $16$ to $256$ in the left and right panels of \ref{fig:dnn-batch}, respectively. Each panel again contains $F_{S_k}(w_k)$, step length, $\hat M_{V,k}$ and $\hat \rho_k$, from top to bottom.
\begin{figure}[h!]
    \centering 
    \includegraphics[width = 0.45\textwidth]{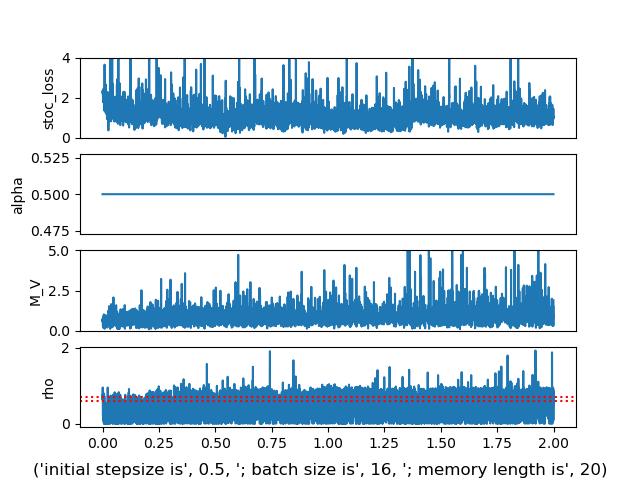}
    \includegraphics[width = 0.45\textwidth]{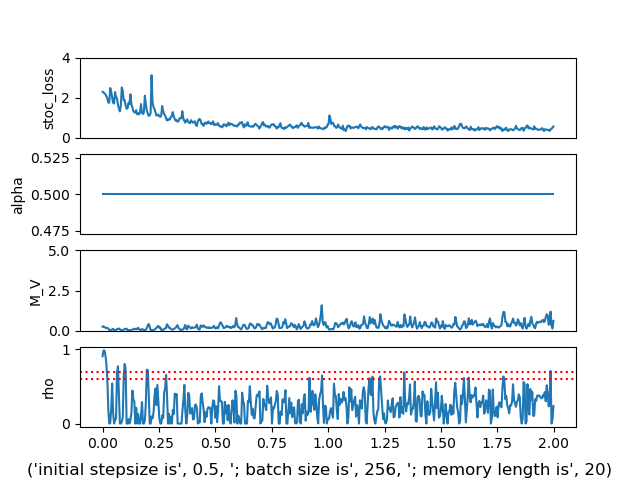}
    \caption{DNN with the Fashion-MNIST data. Each of left and right panels contains a typical run with a constant step length of 0.5; each panel contains from top to bottom: $F_{S_k}(w_k)$, $\hat M_{V,k}$, $\mathbb{V}_k$, $\hat \rho_k(\alpha)$; the left panel has a minibatch size of 16 while the right panel has minibatch size of 256.}
    \label{fig:dnn-batch}
\end{figure}

Similar to Figure \ref{fig:logi_batch}, for deep neural networks results presented in Figure \ref{fig:dnn-batch}, throughout the training progress, when batch size increased from 16 to 256, $\hat M_V$ values reduced, as expected. %the values of $\hat \rho_k$ remained largely comparable 

\subsection{Step size's influence on $\hat \rho_k$}

As discussed in the section around (\ref{eq:hatrho}), we expect $\hat \rho_k$ to be close to $1$ when step lengths are picked to be small (as analytical functions are locally approximated by linear ones) and may oscillate wildly when step lengths are large relatively---we demonstrate this effect in this subsection.

In the logistic regression task, we fixed a batch size of $8$ and varied the step length from $0.3$ to $3e-4$ and examined the resulting values of $\hat\rho_k(0.3)$ and $\hat\rho_k(3e-4)$ in Figure \ref{fig:Logi-step}. We observed that the values of $\hat\rho_k(0.3)$ were widely dispersed across the interval [0,1], with many values being close to 0, indicating potentially too large of step sizes. In contrast, the values of $\hat\rho_k(3e-4)$ were clustered around 1, indicating more progress maybe available for larger step sizes.

%\shig{[Based on these two plots, it can be observed that the value of $\hat M_{V,k}$ remained largely unchanged across various batch sizes and exhibited a slight upward trend during the training process. This behavior is consistent with the fact that $\hat M_{V,k}$ serves as an indicator of the noise level in the gradient. As the algorithm approaches stationarity, the gradient estimates can become more noisy, as demonstrated in one of Bertsakas's example \cite{Bertsaka}.]}

\begin{figure}[h]
    \centering
    \includegraphics[width = 0.49\textwidth]{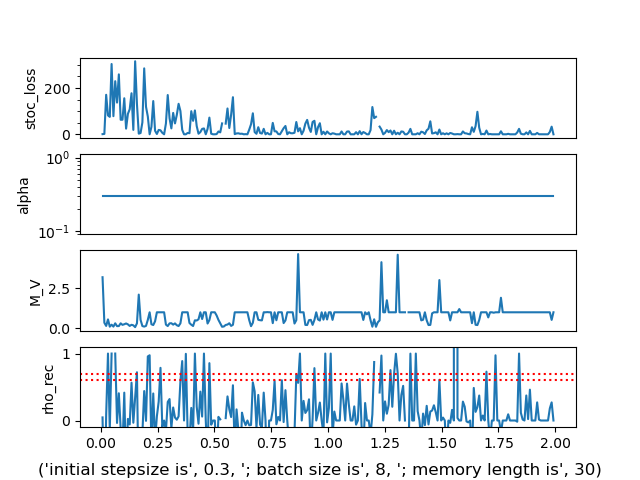}
    \includegraphics[width = 0.49\textwidth]{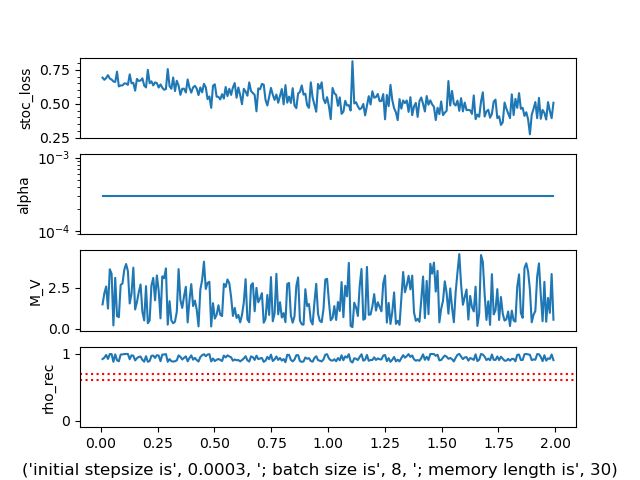}
    \caption{Logistic Regression with Different Step Lengths. Left column: Steplength 3e-1; Right column: Steplength 3e-4.}
    \label{fig:Logi-step}
\end{figure}

We conducted similar experiments on DNN for the Fashion-MNIST dataset and varied the step lengths $0.5$ to $5e-4$ and examined the resulting values of $\hat\rho_k(0.5)$ and $\hat\rho_k(5e-4)$ in Figure \ref{fig:DNN-step}. We observed that, similar to logistic regression, $\hat\rho_k(0.5)$ exhibited wild oscillations with values approaching 0, while $\hat\rho_k(5e-4)$ clustered around 1. For this application, a step length of 0.5 improved the stochastic objective but caused erratic oscillations, while 5e-4 resulted in slow training progress or "stalling". SRT would decrease the step length in the former case to prevent oscillations and increase it in the latter case to promote progress.

\begin{figure}[h]
    \centering
    \includegraphics[width = 0.49\textwidth]{figures/MR_16a_0.5c_0.7epo_2_const}
    \includegraphics[width = 0.49\textwidth]{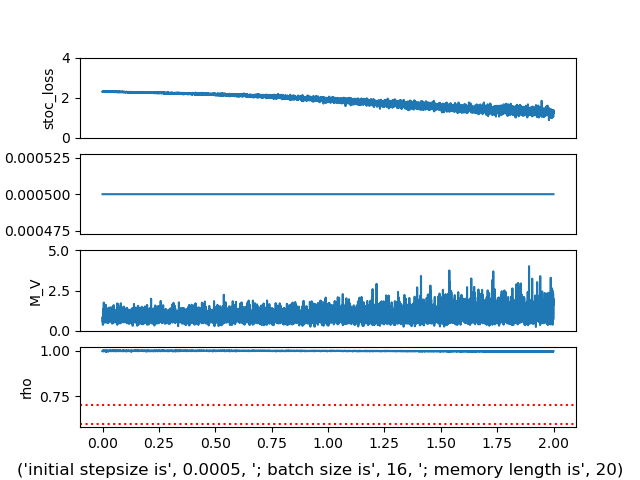}
    \caption{Deep Neural Network with Different Step Lengths. Left column: Step length 5e-1 (erratic oscillation in objective); Right column: Step length 5e-4 (training stalling).}
    \label{fig:DNN-step}
\end{figure}

%From these two plots one observe that the value of $\hat M_{V,k}$ remained largely unchanged across different batch-sizes and has a small upward trend during training, as is expected since this is a indicator of the level of noise in the gradient, and as the iterates approaches stationarity, the gradient estimates can become more noisy (cite bertsakas's example). 

Given that the behaviors of $\hat \rho_k$ and $\hat M_{V,k}$ generally align with our expectations, we will now proceed to present the convergence results for this algorithm in the next section, followed by the application of the algorithm for training tasks in the subsequent section.

\section{Convergence Result for the SRT Algorithm}
We present the main convergence result for the algorithm below and defer its proof to the appendix. 

\paragraph{Convergence Theorem}\textit{
Assume that the stochastic objective function $F_{S_k}(w)$ and the true objective function $F(w)$ are $\mu,L$ smooth; that
$\cE_{\xi_k}[\nabla F_{S_k}\left(w_k\right)] = \nabla F(w_k),$ 
and
$\cE_{\xi_k}[\| \nabla F_{S_k}\left(w_k\right) \|^2]-\|\nabla F_k\|^2\leq M_V\|\nabla F_k\|^2 + M$
and let $c_1$ and $c_2$ be picked such that:
\be 1-\frac{\mu}{2L} < c_1 < c_2  < 1-\frac{\mu}{2\tau L } \ee
and that the estimated $\hat M_{V,k} = M_V$, then 
for all $k$ such that $k > k_0$, where 
$$k_0 = N \cdot \max\left(0 ,\log_\tau\frac{\mu}{\tau L^2 \alpha_0} ,\log_\tau\frac{\alpha_0}{L}\right)$$
the expected optimality gap satisfies the following inequality:
\be \mathbb{E} [F(w_{k}) - F^*]  \leq \eta + \left(1-\frac{2(1-c_1)\mu}{\tau(M_V+1)L}\right)^{k-k_0}  \left[F(w_{k_0}) - F^* -\eta\right]
\ee
where
\be \eta = \frac{\tau M}{4\mu(1-c_1)(M_V+1)}.\ee
}

%\shig{This theorem highlights that under $\mu-L$ smoothness assumptions of the stochastic and true objective, the proposed algorithm is convergent to a stationary neighborhood of the true objective measured by the size $\eta$, which is proportional to the non-diminishing error term $M$ in the second movement bound of the gradient. Furthermore, after the optimal step length is reached at iteration $k_0$, the algorithm also enjoys a linear convergence rate that depend on the conditioning of the problem.}

This theorem highlights that under $\mu, L$ smoothness assumptions of the stochastic and true objective functions, the algorithm proposed is convergent towards a stationary neighborhood surrounding the true objective of size $\eta$. The stationary neighborhood $\eta$ is proportional to the non-diminishing error term denoted as $M$ in the second movement bound of the gradient. The theorem further shows that the optimal step length is attained at iteration $k_0$, after which the algorithm additionally manifests a linear convergence rate that depends upon the problem's conditioning. 

While the algorithms enjoy a favorable linear convergence rate to the stationary neighborhood as characterized by the optimality gap, the somewhat restrictive assumptions leave room for future research directions. For instance, we have assumed for simplicity that the estimated $\hat M_V$ reflects the true value. As a potential future direction, one may assume the distribution of $\|\grad F_{S_k}\|$ and derive high probability bounds for the estimated values. Another limitation of the analysis is that the choices of $c_1,c_2$ require information about the conditioning of the problem. While many practical methods exist for estimating this information, we have found in practice that the algorithm is robust with respect to different choices of $c_1,c_2$, and that this restriction is purely technical to ensure that the step length eventually settles to a fixed value. % rather than diverging or oscillating

\section{Numerical Experiments with SRT}
\label{headings}

To show SRT algorithm's capability of automatic step length tuning, we tested with logistic regression and deep neural network training tasks. All codes were written in Python; experiments on Deep Neural Networks were implemented in Pytorch with BackPACK\cite{dangel2020backpack}. Logistic regression experiments are done on a MacBook Pro with Intel i7 processor with 32 GBs of DDR4 RAM. Deep Neural Network experiments were done on a PC with Nvidia GeForce GPU with 11GB of dedicated VRAM. 

%and we observed that when the step size is intentionally initialized to be too large or small, the SRT algorithm can automatically adjust it to be an appropriate step length. The step lengths obtained from the SRT algorithm are comparable to our hand-tuned step lengths for both tasks.

For the first set of experiments reported in Figure \ref{fig:logidiff}, we ran logistic regression on the gisette dataset for 10 epochs with a batch size of $8$, and initialized the step length at $0.1$ and $1e-4$, in the left and right panel respectively. We recorded the stochastic objective value in the top panels, the adjusted step lengths $\alpha$ in the middle panels, and the computed $\hat \rho$ in the bottom panels.

\begin{figure}[h]
    \centering 
    \includegraphics[width = .49\textwidth]{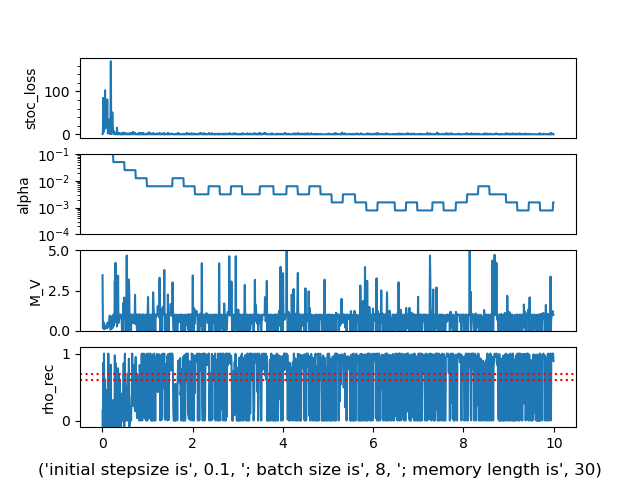}
    \includegraphics[width = .49\textwidth]{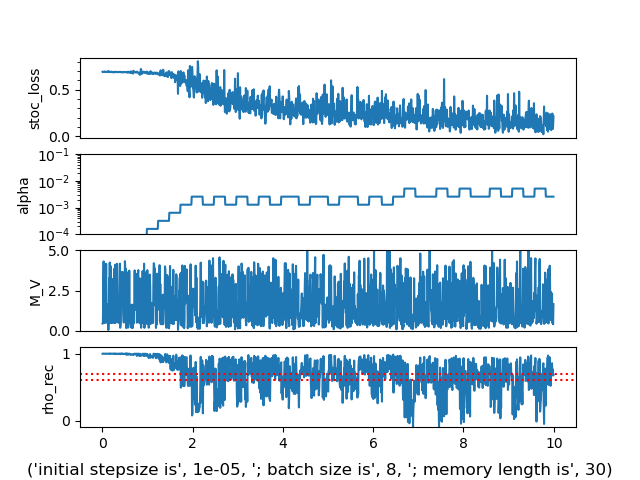}
    \caption{SRT on Logistic Regression with Different Initial Step Lengths. Left column:Initial Steplength 1e-1; Right column: Initial Steplength 1e-5; Each of the panels contains from top to bottom: $F_{S_k}(w_k)$, $\hat M_{V,k}$, $\mathbb{V}_k$, $\hat \rho_k(\alpha)$.}
    \label{fig:logidiff}
\end{figure}

As observed in Figure \ref{fig:logidiff}, the SRT algorithm identifies step lengths as too large or small and adjusts them accordingly. In both cases, the step lengths settled slightly above 1e-3, which match our best hand-tuned step length.

For DNN training tasks, the behavior observed in Figure \ref{fig:dnndiff} is comparable to what was observed before, where we applied the SRT algorithm to train a 3-layer feed-forward neural network on the Fashion-Mnist dataset for 20 epochs. The SRT algorithm reduces the step length when it is too large and increases it when it is too small. And in both cases, the final step length settled to around 0.06, again comparable to our best hand-tuned step length. 
\begin{figure}[h]
    \centering 
    \includegraphics[width = .49\textwidth]{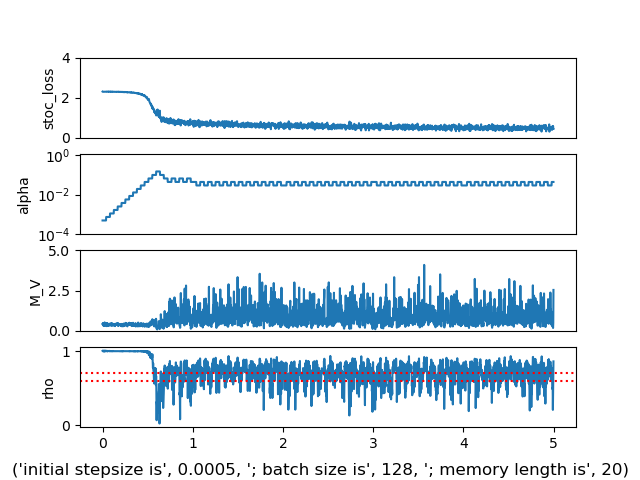}
    \includegraphics[width = .49\textwidth]{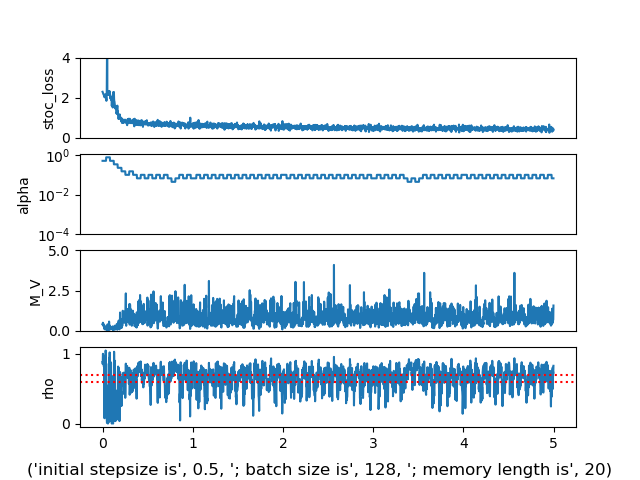}
    \caption{SRT on the training of a feed-forward 3-layer Neural Network on the Fashion-Mnist data with batch size 128. Left column:Initial Steplength 0.5; Right column: Initial Steplength 0.0005. Each of the panels contains from top to bottom: $F_{S_k}(w_k)$, $\hat M_{V,k}$, $\mathbb{V}_k$, $\hat \rho_k(\alpha)$.}
    \label{fig:dnndiff}
\end{figure}

{In both Logistic regression and DNN training tasks, with initially too short or too long of step length choices, SRT algorithm was successful in identifying and adjusting to step lengths which eventually settle into similar values that are comparable to the best hand-tuned values.} %\yx{\sout{In practice, we have attempted different choices of $c_0$ and $c_1$ as well as other statistical tests and establishing confidence intervals rather than using only the average values of $\hat\rho$, and the results were similar.}(Suggest to remove, otherwise will be asked to provide data.)}

%\section{Future Directions}

%\shig{The estimation of $\hat M_{V,k}$ is currently at an early stage of development, indicating that further advancements and refinements are required. Similarly, the evaluation of $\hat \rho_k$ can be conducted with enhanced sophistication, suggesting the potential for more advanced hypothesis testing techniques. Additionally, it is desirable to analyze the algorithm with a focus on high probability outcomes, thereby providing a more rigorous assessment. Furthermore, while this paper primarily considers the basic mini batch gradient direction, it is worth noting that alternative directions can be explored and incorporated into future investigations.}

\section{Conclusions}
This paper presents a novel step length selection algorithm, namely SRT, which stands out from traditional methods by obviating the need for manual tuning efforts and enabling automatic step size adjustments. The proposed algorithm is shown to be convergent under mild assumptions, and the numerical results demonstrate its competitive performance across different training tasks. In particular, SRT produces effective step lengths that are comparable to those obtained from manual tunning. 

As the proposed estimations of $\hat \rho_k$ and $\hat M_{V,k}$ offer novel insights for step length tuning and are applicable to various problems and tasks, future work may involve conducting numerical experiments with acceleration directions instead of the steepest descent direction. Additionally, relaxing the assumptions necessary for convergence analysis by incorporating the underlying statistical nature could be explored. We leave these potential avenues for future research.

\begin{ack}
Sun was supported by NSF grant DMS-1620022. %Nocedal was supported by AFOSR grant FA95502110084 and ONR grant N00014-21-1-2675.

The authors would like to acknowledge Prof.Jorge Nocedal for his perceptive insights and critical evaluations of this work, which greatly enriched the theory and interpretation presented in this paper. 

The authors would also like to thank Prof.Richard Byrd and Prof.Figen \"Oztoprak for their feedback.
\end{ack}
\newpage

\bibliographystyle{plain}

\bibliography{references}

\newpage
\section*{Appendix}
In this section we present the proof for the main global convergence theorem:
\paragraph{Convergence Theorem}\textit{
Assume that the stochastic objective function $F_{S_k}(w)$ and the true objective function $F(w)$ are $\mu,L$ smooth; that
$\cE_{\xi_k}[\nabla F_{S_k}\left(w_k\right)] = \nabla F(w_k),$ 
and
$\cE_{\xi_k}[\| \nabla F_{S_k}\left(w_k\right) \|^2]-\|\nabla F_k\|^2\leq M_V\|\nabla F_k\|^2 + M$
and let $c_1$ and $c_2$ be picked such that:
\be 1-\frac{\mu}{2L} < c_1 < c_2  < 1-\frac{\mu}{2\tau L } \ee
and that the estimated $\hat M_{V,k} = M_V$, then
for all $k$ such that $k > k_0$, where
$$k_0 = N* \max\left(0 ,\log_\tau\frac{\mu}{\tau L^2 \alpha_0} ,\log_\tau\frac{\alpha_0}{L}\right)$$
the expected optimality gap satisfies the following inequality:
\be 
\mathbb{E} [F(w_{k}) - F^*]  \leq \eta + \left(1-\frac{2(1-c_1)\mu}{\tau(M_V+1)L}\right)^{k-k_0}  \left[F(w_{k_0}) - F^* -\eta\right]
\ee
where
\be \eta = \frac{\tau M}{4\mu(1-c_1)(M_V+1)}.\ee
}

\textbf{Proof}: For any particular iteration $k$, by $\mu,L$ smoothness assumption of $F_{S_k}$, 
\be
\alpha \nabla F_{S_k}^Td_k+\frac{\alpha^2\mu}{2} \left\|d_k\right\|_2^2 
\leq 
F_{S_k}( w_k + \alpha d_k) - F_{S_k}(w_k)  
\leq 
\alpha \nabla F_{S_k}^Td_k+\frac{\alpha^2 L}{2} \left\|d_k\right\|_2^2 
\ee
using the fact that $d_k = \grad F_{S_k}$,
\be
-\alpha \|\nabla F_{S_k}\|^2+\frac{\alpha^2\mu}{2} \left\|\nabla F_{S_k}\right\|_2^2 
\leq 
F_{S_k}( w_k + \alpha d_k) - F_{S_k}(w_k)  
\leq 
-\alpha \|\nabla F_{S_k}\|^2+\frac{\alpha^2L}{2} \left\|\nabla F_{S_k}\right\|_2^2 
\ee
dividing this inequality by $-\alpha \|\nabla F_{S_k}\|^2$ and combining terms:
\be 1 - \frac12 \mu\alpha\geq \frac{F_{S_k}( w_k + \alpha d_k) - F_{S_k}(w_k)  }{-\alpha \|g_k\|^2} \geq 1 - \frac12 L\alpha\ee
using definition of $\hat \rho_k(\alpha)$ in \eqref{eq:hatrho} this implies:
\be \hat \rho_k(\alpha) \in \left[ 1 - \frac{L\alpha}2 , 1 - \frac{\mu\alpha}2 \right]\ee

If $\alpha > 1/L$ at iteration $k$ this implies that 
\be \hat \rho_k(\alpha) \leq 1 - \frac{\mu\alpha}2 < 1-\frac{\mu}{2L}< c_1\ee
Since the same argument for $k$ will apply for $k+1, ... , k+N-1$, $\alpha$ will be decreased based on the algorithm within $N$ iterations.% \shig{[will make this precise in publication]}

Conversely, if $\alpha < \frac1{\tau L} \frac\mu{L}$ at iteration $k$ we have 
\be \hat \rho_k(\alpha) \geq 1 - \frac{L\alpha}2 > 1-\frac{\mu}{2\tau L } > c_2\ee
Since the same argument for $k$ will apply for $k+1, ... , k+N-1$, $\alpha$ will be increased based on the algorithm within $N$ iterations.% \shig{[will make this precise in publication]}

Thus after finitely many iterations, 
\be \alpha_k \in \left(  \frac1{\tau L} \frac\mu{L}  , \frac1L\right) \ee
Also note: $\frac{\mu}{L} > 2(1-c_1)$ by the choice of $c_1$,
\be \alpha_k \in \left(  \frac1{\tau L} \frac\mu{L}  , \frac1L\right) \subset \left(  \frac{2(1-c_1)}{\tau L}  , \frac1L\right). \ee
Since 
\be \frac1L \left/  \left( \frac1{\tau L} \ \frac\mu{L}\right) \right.  > \tau, \ee
$\alpha_k$ must eventually settle to a value in this interval. Furthermore, the number of iterations required to produce a step in the above interval is 
\be k_0 = N* \max\left(0 ,\log_\tau\frac{\mu}{\tau L^2 \alpha_0} ,\log_\tau\frac{\alpha_0}{L}\right)\ee
Now consider the case when $k > k_0$  i.e. a final step size $\bar \alpha$ is achieved such that 
\be\bar \alpha \in  \left(  \frac{2(1-c_1)}{\tau L}  , \frac1L\right)\ee

%The remainder of the proof follows the fixed step length analysis (to be typed up/can directly cite theorem in the review paper).
From this point on, we assume that the optimal step size $\bar\alpha$ is achieve, by the assumption that $\hat M_V = M_V$, the steps taken will admit:
\be w_{k+1} = w_{k} - \frac{\bar\alpha}{M_V+1}\grad F_{S_k}(w_k)\ee
By the $L-$Lipschitz continuity of $F$, we have
\be
F\left(w_{k+1}\right)-F\left(w_k\right) \leq \nabla F\left(w_k\right)^T\left(w_{k+1}-w_k\right)+\frac{1}{2} L\left\|w_{k+1}-w_k\right\|_2^2 
\ee
\be
 \leq-\frac{\bar\alpha}{M_V+1} \nabla F\left(w_k\right)^T \nabla F_{S_k}\left(w_k\right)+\frac{1}{2} \left(\frac{\bar\alpha}{M_V+1}\right)^2 L\left\|\nabla F_{S_k}\left(w_k\right)\right\|_2^2
\ee
Taking expectation with respect to $\xi_k$, we obtain
\be
F\left(w_{k+1}\right)-F\left(w_k\right)\leq-\frac{\bar\alpha}{M_V+1} \| \nabla F\left(w_k\right)\|^2+\frac{1}{2} \left(\frac{\bar\alpha}{M_V+1}\right)^2 L\cE_{\xi_k}[\left\|\nabla F_{S_k}\left(w_k\right)\right\|_2^2]
\ee
By the assumption that
$\cE_{\xi_k}[\| \nabla F_{S_k}\left(w_k\right) \|^2]-\|\nabla F(w_k)\|^2\leq M_V\|\nabla F(w_k)\|^2 + M$,
we have
\be\cE_{\xi_k}[\| \nabla F_{S_k}\left(w_k\right) \|^2]\leq (M_V+1)\|\nabla F(w_k)\|^2 + M\ee
Plug this into the previous equation and obtain:
%\be
%F\left(w_{k+1}\right)-F\left(w_k\right)\leq-\frac{\bar\alpha}{M_V} \|  \nabla F\left(w_k\right)\|^2+\frac{1}{2} \left(\frac{\bar\alpha}{M_V}\right)^2 L\left[ (M_V+1)\|\nabla F(w_k)\|^2 + M\right]
%\ee

\be
\begin{split}
F(w_{k+1}) - F(w_k) &\leq -\frac{\bar{\alpha}}{M_V+1} \left\| \nabla F(w_k) \right\|^2 + \frac{1}{2} \left( \frac{\bar{\alpha}}{M_V+1} \right)^2 L \left[ (M_V + 1) \left\| \nabla F(w_k) \right\|^2 + M \right] \\
%&= -\frac{\bar{\alpha}}{2(M_V+1)} \left\| \nabla F(w_k) \right\|^2 + \frac{\bar{\alpha}^2}{2(M_V+1)^2} L (M_V + 1) \left\| \nabla F(w_k) \right\|^2 + \frac{\bar{\alpha}^2 M}{2(M_V+1)^2} L \\
&= -\frac{\bar{\alpha}}{M_V+1} \left[ 1 - \frac{\bar{\alpha} L }{2} \right] \left\| \nabla F(w_k) \right\|^2 + \frac{\bar{\alpha}^2 L}{2(M_V+1)^2} M \\
&= -\frac{\bar{\alpha}}{2(M_V+1)} \left\| \nabla F(w_k) \right\|^2 + \frac{\bar{\alpha}^2 L}{2(M_V+1)^2} M\\
&\leq -\frac{\bar{\alpha}\mu}{M_V+1}  [F(w_k) - F^*] + \frac{\bar{\alpha}^2 L}{2(M_V+1)^2} M
\end{split}
\ee
rearranging terms, subtracting $F^*$ from both sides and obtain:
\be
\begin{split}
F(w_{k+1}) - F^* &\leq \left(1-\frac{\bar{\alpha}\mu}{M_V+1}\right)  [F(w_k) - F^*] + \frac{\bar{\alpha}^2 L}{2(M_V+1)^2} M\\
&\leq\left(1-\frac{2(1-c_1)\mu}{\tau(M_V+1)L}\right)  [F(w_k) - F^*] + \frac{M}{2L(M_V+1)^2}
\end{split}
\ee
We subtract $\frac{\tau M}{4\mu(1-c_1)(M_V+1)}$ from both sides and obtain
\be
\begin{split}
F(w_{k+1}) - F^* - \frac{\tau M}{4\mu(1-c_1)(M_V+1)} \leq \left(1-\frac{2(1-c_1)\mu}{\tau(M_V+1)L}\right)  \left[F(w_k) - F^* -\frac{\tau M}{4\mu(1-c_1)(M_V+1)}\right]
\end{split}
\ee

Recursively apply this argument from $k_0$ to $k$ to attain the argument.

\end{document}